# Prediction of IPL Match Outcome Using Machine Learning Techniques

Srikantaiah K C[1,*], Aryan Khetan[1], Baibhav Kumar[1], Divy Tolani[1], Harshal Patel[1]

[1]*Department of CSE, SJB Institute of Technology, Affilated to Visveswaraya Technological University BGS Health & Education City, Bengaluru-560060, Karnataka, India*
[*]*Corresponding author. Email: srikantaiahkc@gmail.com,*

**ABSTRACT**
India's most popular sport is cricket and is played across all over the nation in different formats like T20, ODI, and Test. The Indian Premier League (IPL) is a national cricketmatch where players are drawn from regional teams of India, National Team and also from international team. Many factors like live streaming, radio, TV broadcast made this league as popular among cricket fans. The prediction of the outcome of the IPL matches is very important for online traders and sponsors. We can predict the match between two teams based on various factors like team composition, batting and bowling averages of each player in the team, and the team's success in their previous matches, in addition to traditional factors such as toss, venue, and day-night, the probability of winning by batting first at a specified match venue against a specific team. In this paper, we have proposed a model for predicting outcome of the IPL matches using Machine learning Algorithms namely SVM, Random Forest Classifier (RFC), Logistic Regression and K-Nearest Neighbor. Experimental results showed that the Random Forest algorithm outperforms other algorithms with an accuracy of 88.10%.

***Keywords:*** *Cricket, Indian Premier League, Logistic Regression, Machine Learning, Prediction of match outcome, Random Forest Classifier.*

## 1. INTRODUCTION

Cricket is an outdoor game which is played by bat and bowl which includes 2 teams of 11 players each. Cricket is a teamwork game and is played mostly in three formats and occupies the 2 spots in the list of the most popular sport around the World. Like in any sport, there are many factors that plays an importantrole in deciding the winner of the match. Selection of a team is based on the player performance and other considerations like pitch factor, team size, venue etc. There are many variables and constraints which makes The Analysis of Cricket Match Difficult. There are three different formats of Cricket namely - Tests, Twenty-twenty (T20) and One Day International (ODI). Cricket is not only a nation game but also an international game. In this game, every ball is crucial because every ball can change the whole match in Cricket [15, 16].

Indian Premier League (IPL) is a national cricketmatch where players are drawn from regional teams of India, National Team and also from international team. It is based on 20-20 format and is owned by Celebrities, Businessmen and others and the entire IPL is controlled by Board of Control for Cricket in India (BCCI). For the current year (2021)there are total of 8 Teams in IPL namely, Royal Challengers Bangalore (RCB), Rajasthan Royals (RR), Chennai Super Kings (CSK), Mumbai Indians (MI), Kolkata Knight Riders (KKR), Delhi Capitals (DC), Punjab Kings (PK) and SunRisers Hyderabad (SRH). The motivation behind this paper includes the answers to following questions: *"What is the probability of winning the game at a particular venue based on decision to field/bat first on winning the toss?", "Most dismissals by a bowler in a match?", "Does Home Ground have any effect on the result of the game?*

In this paper We are trying to find out the match winner of an IPL match based on the stadium they are choosing and the toss decision using machine learning techniques like SVM, Random Forest, Logistic Regression etc. Remainder of the paper is organized as follows: The section 2 is the literature survey, section 3 deals with the problem definition and the architecture. Section 4 deals with the experimental results. Section 5 talks about the conclusion.

## 2. LITERATURE SURVEY

Ahmad *et al.* [1], predicted the emerging players from batsman as well as from the bowlers using machine learning techniques. Song *et al.* [2] predicted estimation of the location of a moving ball based on the value of the cricket sensor network. Roy *et al.* [3] predicted ranking system which is based on the social network factors and their evaluation in the form of composite distributed framework using Hadoop framework and MapReduce programming model is used for processing the data. Priyanka *et al.* [4], predicted the outcome of IPL-2020 based on the 2008-2019 IPL datasets using Data Mining Algorithms with an accuracy of 82.73%.

Kansal *et al.* [5], predicted player evaluation in IPL based on the 2008-2019 datasets using Data Mining Technique. Data mining algorithms are used which gives evaluation using player statistics assessing a player's performance and determining his base price. They predicted about how to select a player in the IPL, based on every player's performance history using algorithms like decision tree, Naïve Bayes and Multilayer perceptron (MLP). MLP outperforms better than other algorithms. Agrawal *et al.* [6], used Support Vector Machine (SVM), CTree, and Naïve Baiyes classifiers with accuracies of 95.96%, 97.97% and 98.98% respectively, to predict the probability of the winner of the matches. Barot *et al.* [7], predicted the match outcome based on the toss and venue.

Kaluarachchi *et al.* [8], predicted match outcome using home ground, time of the match, match type, winning the toss and then batting first by using Naïve Bayes classifier. Passi *et al.* [9], predicted the performance of players based on the runs and the number of wickets. Both the type of problems is treated as classification problems where the list of runs, and list of wickets are classified in different ranges based on machine learning algorithms. The Random Forest algorithm outperforms better than other algorithms. Nigel Rodrigues *et al.* [10], predicted the value of the traits of the batsmen and the bowlers in the current match. This would help in selecting the players for the upcoming matches by using past performances of a player against a specific opposition team by using Multiple Random Forest Regression.

Wright [11], predicted the possible fixture for a cricket match based on the various venue, teams, number of holidays between each match in a fair and efficient manner. A metaheuristic procedure is used to progress from the basic solution to a complex final solution by a technique, Subcost-Guided Simulated Annealing (SGSA). Maduranga *et al.* [12], predicted the outcome of any cricket match by using data mining algorithms and provided solutions for the approach used by other authors. Shetty *et al.* [13], predicted the capabilities of each player depending on various factors like the ground, pitch type, opposition team and several others by using machine learning techniques. The model gave an accuracy of 76%, 67%, and 96% for batsmen, bowlers, and all-rounders respectively by using Random Forest Algorithm. This model helped them to select the best players of the game and predict outcomes of the match [25-29].

## 3. PROBLEM DEFINITION AND ARCHITECTURE

### *3.1. Problem Definition*

Given IPL datasets of past 9 years, the main objective of this paper is to predict the outcome of an IPL match between two teams based on the analysis of previously stored data using Machine Learning algorithms. The information will be analyzed and preprocessed. After preprocessing the data will be used to train different models in order to give the outcomes. We will analyze the various datasets and use key variables such as strike rate, bowler economy, etc. and feed it as input to an algorithm will help us get the probable outcome of a match [16-19].

### *3.2. Architecture*

The figure 1 represents the architecture of the model which includes different components like datasets, split data, Training, Testing, Supervised Learning models and Result.

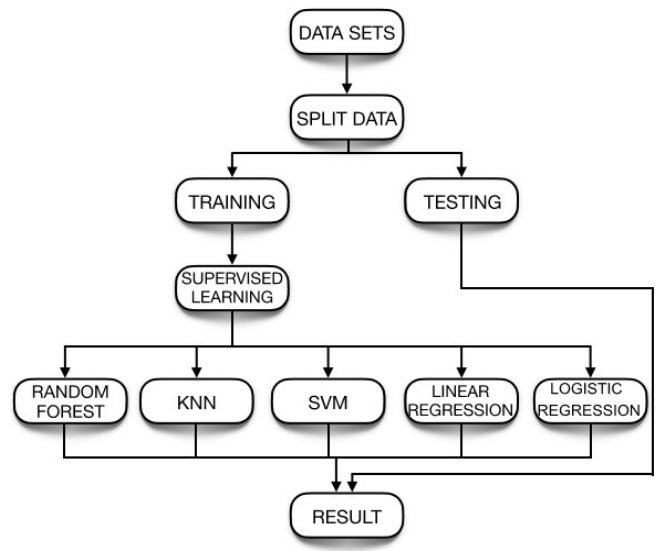

**Figure 1** Architecture of the model

### *3.2.1. Dataset*

The first step in the Architecture of model is to collect datasets from various sources. The data which is fed into the model decides how the model acts and reacts. If the data is accurate and up-to-date, then we will have accurate outcomes or predictions. So, we have collected 6 datasets from Kaggle.com which are as follows [20-21].

*3.2.1.1. Team wise home and away dataset*

The Teamwise Home and Away dataset contains 6 columns for the datasets which are as follows: home_wins, away_wins, home_matches, away_matches, home_win_percentage and away_win_percentage. It describes about the team performance in the home and away conditions with their win percentage. The table 1 shows the dataset and its description [22].

**Table 1.** Teamwise Home and Away dataset description

| Column name | Column description |
|---|---|
| Home_wins | Tells the number of matches won by a team in their home ground. |
| away_wins | Tells the number of matches won by a team other than their home ground. |
| home_matches | Tells the number of matches played by a team in their home ground. |
| Away_matches | Tells the number of matches played by a team other than their home ground. |
| home_win_percentage | Percentage of matches won when played in home ground. |
| Away_win_percentage | Percentage of matches won when played in ground other than their home ground. |

*3.2.1.2. Matches dataset*

The Matches datasets contains 16 columns i.e season, city, date, team1, team2, toss_winer, toss_decision, result, dl_applied, winer, win_by_runs, win_by_matches and player_of_the_match, venue, umpire1 and umpire2. This dataset tells about the matches that are played between two teams and who was the winner of the match. It also tells about the toss decision taken in the match. The table 2 shows the dataset column and its description [23].

**Table 2.** Matches dataset description

| Column name | Column description |
|---|---|
| Season | This column tells the season in which the match was played. |
| City | The city in which the match was played. |
| Date | The date on which the match was played. |
| Team 1 | Name of team who played match. |
| Team 2 | Name of team which played match. |
| Toss winner | Team that won the toss. |
| Toss decision | Decision of batting or fielding after winning toss. |
| Result | Outcome of match whether normal or tie. |
| Dl_applied | Information on whether DL method was applied or not. |
| Winner | Team that won the match. |
| Win by runs | The number of runs by which the team won. |
| Win by wickets | The number of wickets by which the team won. |
| Player of match | Name of player who was awarded player of the match. |
| Venue | Stadium in which match took place. |
| Umpire 1 | On field umpire name. |
| Umpire 2 | On field umpire name. |

*3.2.1.3. Player's dataset*

The Player's dataset contains 5 columns namely Player_Name, DOB, Batting_Hand, Bowling_Skill and Country. This dataset tells about the player and his bowling and batting style. The table 3 shows the dataset column and its description.

**Table 3.** Player's dataset description

| Column name | Column description |
|---|---|
| Player Name | Name of the players |
| DOB | Date of birth of the players |
| Batting Hand | Tells whether the players are left-handed or right-handed. |
| Bowling skill | Bowling style of players |
| Country | Name of countries to which the player belongs. |

### *3.2.1.4. Teams Datasets*

The teams' datasets contain a single column named as team1 which shows the various IPL teams. The table 4 shows the dataset column and its description.

**Table 4.** Teams dataset description

| Column Name | Column Description |
|---|---|
| Team | Names of teams participating in the IPL tournament. |

### *3.2.1.5. Deliveries dataset*

The Deliveries dataset contains 20 columns i.e innings, batting_team, bowling_team, over, Ball, Batsmen, Non_striker, Bowler, is_super_over, wide run, bye_run, Legbye_run, no_ball_run, penalty_run, batsmen_run, extra_runs, total_runs, players_dismissed, dismissal_kind, fielder. The table 5 shows the dataset column and its description [24].

**Table 5.** Deliveries dataset description

| Column Name | Column Description |
|---|---|
| Inning | Tells us the inning being played |
| Batting_team | Tells us name of then batting team |
| Bowling_team | Tells name of bowling team |
| Over | Tells the over number being bowled |
| Ball | Tells the ball number of the over |
| Batsmen | Name of batsmen on strike |
| Non-striker | Name of batsmen on runner end |
| Bowler | Name of bowler bowling |
| Is_super_over | Tells if over is super over or not |
| wide_run | If there are runs given for wide ball |
| bye_run | If there are any bye runs given |
| Legbye_run | If there are any leggy runs given |
| no_ball_run | If the ball was a no ball |
| penalty_run | Penalty runs due to any reason |
| batsmen_runs | Runs hit by batsmen on the ball |
| extra_runs | Total extra runs given |
| total_runs | Total runs in the ball |
| player_dismissed | If the player was given out or not |
| dismissal_kind | What kind of dismissal it was |
| Fielder | Player who caused dismissal |

### *3.2.1.6. Most_runs_average_strikerate dataset*

The Most_runs_average_strikerate dataset contains 6 columns namely batsman, total_runs, out, numberofballs, average, strikerate. This tells about player batting statistics. The table 6 shows the dataset column and its description.

**Table 6.** Most_runs_average_strikerate dataset description

| Column name | Column description |
|---|---|
| Batsman | Name of the batsman. |
| Total_runs | Total runs scored by the batsman |
| Out | Tells about the number of times the batsman got out in the entire IPL career. |
| No_of_balls | Number of balls faced by the batsman overall. |
| Average | The average runs of the batsman |
| Strikerate | Tells us about the strike rate of the batsman. |

### *3.2.2. Split Data*

In this step the dataset is splitted into two groups, one for testing and one for training. The training data is used to train the Machine learning algorithms using supervised learning techniques. The trained model is then tested using the algorithms and the result is predicted. The testing data and training data are divided in the ratio 30:70.

### *3.2.3. Training the model*

Training is the most important stage in Machine Learning. In this step the model is trained using training data to find patterns and make predictions. It results in the model learning from the dataset so that it can accomplish the given task.

### *3.2.4. Testing the model*

After training the model, the performance of the model is checked. This is done by testing the model with previously unseen datasets. The unseen Datasets is called the testing datasets. So, in this step the model is tested by providing the unseen testing data.

### 3.2.5. Supervised Learning

There are different types of Supervised Machine Learning Techniques like Logistic Regression, Support Vector Machine, K-Nearest Neighbor, Random Forest classifier, linear Regression etc.

## 4. EXPERIMENTAL RESULTS

### 4.1. Experimental Setup

The Experiments were carried out on minimum Hardware requirements that include Processor of Intel Atom or Intel core i3, Disk space should be at least 1GB, Operating System must be windows 7 or later, macOS and Linux. The machine learning algorithm was implemented using Python 3.8 and Jupyter Notebook. The Datasets were downloaded from Kaggle.

### 4.2. Analysis of IPL datasets

The IPL dataset was analyzed, and some features were extracted out such as team winning percentage while batting first and team winning percentage while bowling first which is shown in figure 2 and figure 3.

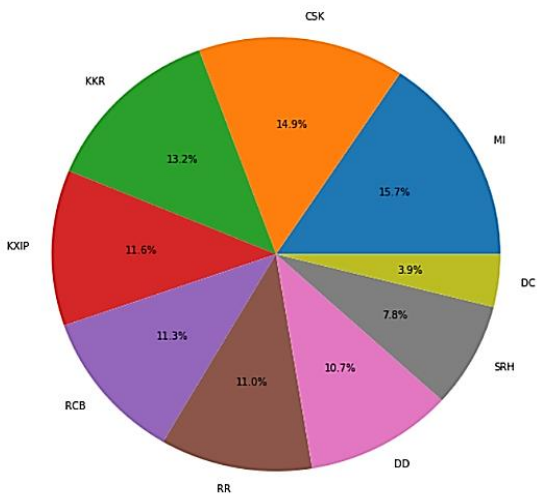

**Figure 2** Team winning percentage while batting first

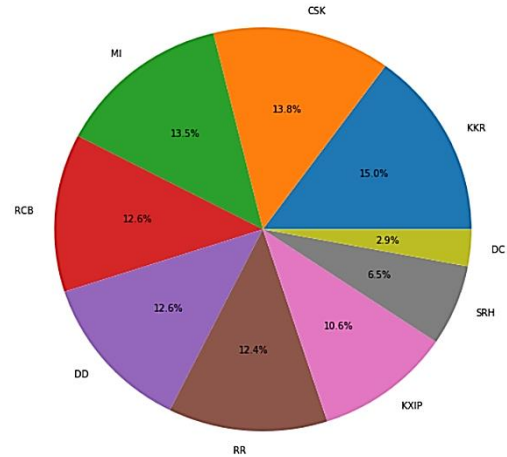

**Figure 3** Team winning percentage while bowling first

Away win percentage is also being displayed in figure 4 below. From the figure 4, it can be seen that GL has the highest away win percentage.

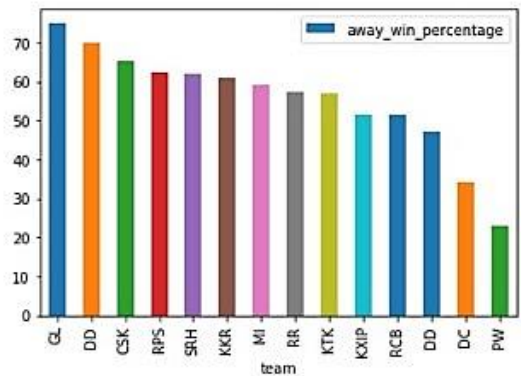

**Figure 4** Teamwise winning percentage in away matches

Home win percentage is also being displayed in fig 5 below. From the fig 5, it can be seen that RPS has the highest home win percentage.

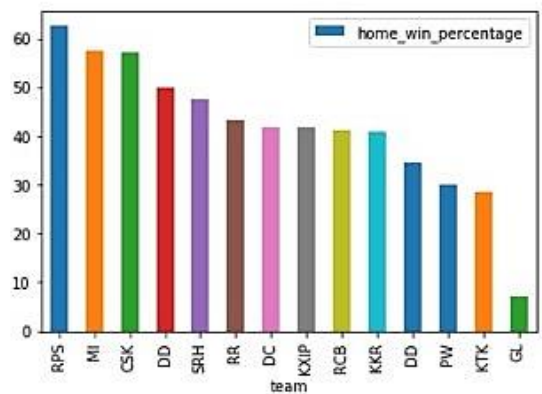

**Figure 5** Teamwise winning percentage in home matches

The fig 6 gives the information about how many times a particular team has won the toss and a pie chart is made from the dataset information depicting the winning percentage of both batting first and batting second at a particular venue after winning the toss in fig 7.

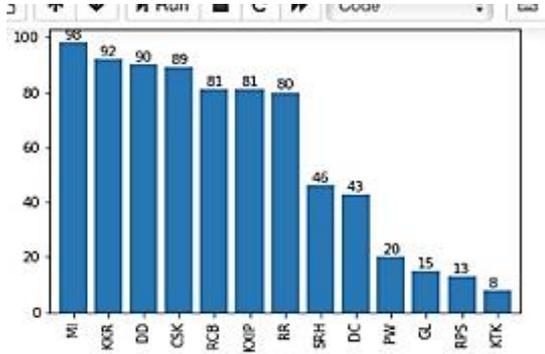

**Figure 6** how many times a particular team won the toss

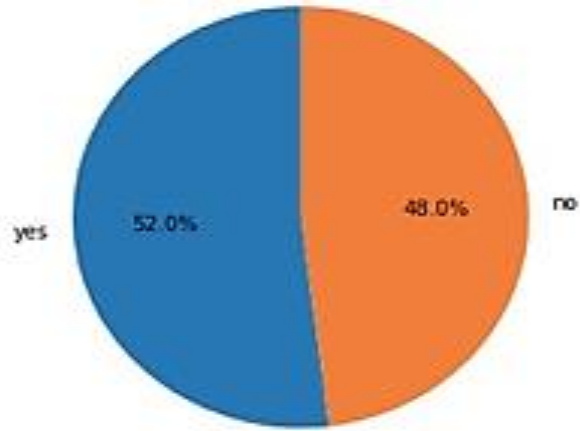

**Figure 7** Winning percentage after winning the toss

The comparison between the two particular teams playing the match is shown through graph which could help in predicting the winner. Comparison between MI and CSK over various seasons is shown over here through figure 8.

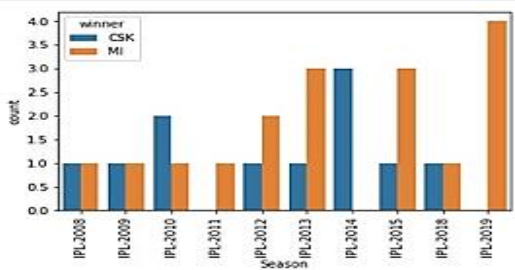

**Figure 8** Team Comparison

Figure 9 displays the highest run scorers of IPL. From the fig, it can be seen that Virat Kohli is the highest run scorer with 5434 runs.

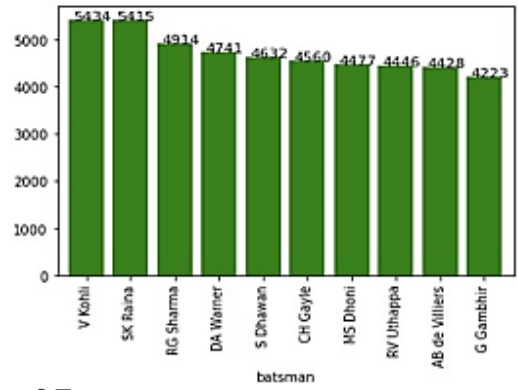

**Figure 9** Top run scorers

## 4.3. Performance Evaluation

1) Accuracy: The fraction of correct forecast in all predictions is known as accuracy. In this experiment, random forest classifier outruns all the algorithms by predicting the result with highest accuracy of 88.10%. The figure 10 shows the accuracies of the various algorithms implemented.

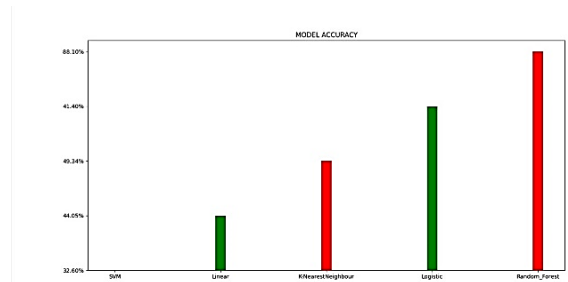

**Figure 10** Accuracy of various algorithms

Table 7 shows about the various algorithms and their accuracies obtained. It is clear from the table that the random forest classifier performed better than other algorithm.

**Table 7.** Accuracy achieved by the algorithms

| Algorithm | Accuracy |
|---|---|
| Random Forest | 88.10% |
| K-Nearest Neighbor | 49.34% |
| Logistic Regression | 51.40% |
| SVM | 32.6% |
| Linear Regression | 44.05% |

Furthermore, 2-fold, 5-fold, 10-fold cross validation for rfc is also implemented for having better insights in table 8

**Table 8.** Cross validation Technique

| NO. OF FOLDS | TRAIN SET | TEST SET | ACCURACY |
|---|---|---|---|
| FOLD 1 | 680 | 76 | 50.0 % |
| FOLD 2 | 680 | 76 | 47 % |
| FOLD 3 | 680 | 76 | 51 % |
| FOLD 4 | 680 | 76 | 58 % |
| FOLD 5 | 680 | 76 | 46 % |
| FOLD 6 | 680 | 76 | 49 % |
| FOLD 7 | 681 | 75 | 55 % |
| FOLD 8 | 681 | 75 | 43 % |
| FOLD 9 | 681 | 75 | 47 % |
| FOLD 10 | 681 | 75 | 53 % |

MEAN VALUE FOR ABOVE K-FOLD IS 50 %

## 5. CONCLUSIONS

Predicting a winner in a sport such as cricket is especially challenging and involves very complex processes. But with the introduction of machine learning, this can be made much easier and simpler. In this paper, various factors have been identified that contribute to the results of the Indian Premier League matches. Factors that have a major impact on the outcome of an IPL match include the teams playing, the venue, the city, the toss winner and the toss decision. We have analyzed IPL data sets and predicted game results based on player performance. The methods used in the work to obtain the final test are Logistic regression, Support Vector Machine (SVM), Decision tree, Random Forest classifier and K-nearest neighborhood. Random Forest classification (RFC) outperforms the other algorithm.

As for the scope of the future, the focus can be on each player's performance and evaluate that on a regular basis for the season. His ratings for bowling and batting can also be predicted. There can be a chance to predict the man of the match for the two teams.

## ACKNOWLEDGMENTS

We would like to express our sincere thanks to the Department of CSE, SJBIT for giving us a chance to work on this paper and our institution, SJB Institute of Technology for providing immense support throughout the paper. We would also like to extend our thanks to Visvesvaraya Technological University for providing us the opportunity to work on this paper.